\documentclass[11pt,a4paper]{article}
\usepackage[hyperref]{acl2019}
\usepackage{times}
\usepackage{latexsym}

\usepackage{amsmath,amssymb}
\usepackage{url}
\usepackage{numprint}
\usepackage[utf8]{inputenc}
\usepackage[pdftex]{graphicx}

\aclfinalcopy 

\title{Towards De-identification of Legal Texts}

\author{Diego Garat  and  Dina Wonsever 
\\ Instituto de Computaci\'on, Facultad de Ingenier\'ia, Universidad de la Rep\'ublica 
\\ \tt \{dgarat,wonsever\}@fing.edu.uy }

\begin{document}
	
\maketitle

\begin{abstract}
In many countries, personal information that can be published or  shared between organizations is regulated and, therefore, documents must undergo a process of de-identification to eliminate or obfuscate confidential data. Our work focuses on the de-identification of legal texts, where the goal is to hide the names of the actors involved in a lawsuit without losing the sense of the story. We present a first evaluation on our corpus of NLP tools in tasks such as segmentation, tokenization and recognition of named entities, and we analyze several evaluation measures for our  de-identification task. Results are meager: 84\% of the documents have at least one name not covered by NER tools, something that might lead to the re-identification of  involved names. We conclude that tools must be strongly adapted for processing texts of this particular domain.
\end{abstract}

\section{Introduction}
For publication of data containing personal information, preservation of people's privacy must be considered. For example, this is the case of documents in the domain of health, where the results of clinical studies, medical procedures and detected pathologies are considered confidential information, or in the legal domain, where names of minors of age or primary offenders should be preserved. In fact, in many countries publication of sensitive data is regulated and can lead to legal sanctions \cite{LEY18331, HIPAA96, UE2016-679}. The protection of privacy comes  in conflict with the availability and use of this information by the scientific community or the general public. As a paradigmatic case, we can mention the information on medical treatments where, due to its volume, research can be supported with statistical methods.

In order to publish information with sensitive data, processes of \emph{de-identification}, or \emph{anonymization}, are investigated and developed. The goal of these systems is to remove or obfuscate  data that allow the individualization of people or institutions. This process of de-identification is often done manually: a costly task, both in time and in human resources, and without guarantees of a totally reliable result \cite{SWEENEY96}.

When the data are in relational databases, k-anonymization algorithms are usually applied, while the problem is different when the information is found in a free text or, even, images \cite{BAYARDO05, DOMINGO-FERRER13, GARDNER08, HEATHERLY15, NEWHAUSER14, KARLE17, PATIL17, ZHANG14, Friedrich19, li18, DERNONCOURT16}.   Increasing the difficulty of the problem, sensitive  names of people, companies or hospitals are contained within a narrative that is not always  grammatical sound.

In the latter case, Natural Language Processing (NLP) techniques might be applied for automatic or semi-automatic anonymization of texts. The detection of different entities make the process of de-identification closely related to the tasks of Recognition of Named Entities (NER) and to the resolution of coreferences, although it is not exempt from its own peculiarities \cite{ARAMAKI06}.

\section{De-identification of legal texts}
Our work focuses on the particular problem of  named entity de-identification in legal texts, which, as shown below, adds the additional task of  stringing of coreference chains \cite {DIAS16} to the entity detection problem. To exemplify, let's consider a fragment of a legal sentence\footnote{All the examples of this article come from our corpus; names are replaced by fantasy names to preserve anonymity}:

\begin{quote}
	\label{ej:sent339-2015}{\small Que ninguno de los funcionarios de la dependencia, incluso el Comisario a cargo constataron u observaron algún tipo de conducta fuera de lugar del Sr. \textbf{Juan Pérez}. Que la denuncia fue realizada por la Sra. \textbf{María Rodríguez} que es quien lideraba la Cárcel y ordenaba a las demás (\ldots) Que lo único admitido por el Sr. \textbf{Pérez} es que compró un chip a la Sra. \textbf{Juana Fernández}, pero no por eso se lo puede acusar de abuso y mucho menos de violación. } \\
	{ \tiny That none of the officials of the dependency, including the Commissioner in charge, stated or observed any type of misplaced conduct by Mr. \textbf{Juan Pérez}. That the complaint was made by Mrs.\textbf{María Rodríguez} who leads the jail and ordered the others (\ldots) that the only thing admitted by Mr. \textbf{Pérez} is that he bought a chip from Mrs. \textbf{Juana Fernández}, but that does not mean he can be accused of abuse and much less of rape.}
\end{quote}

For the de-identification of named entities, the simplest solution consists of completely eliminating the information, replacing it by some generic label like "X", similar to striking out names on a printed document. This procedure, although effective for hiding names, does not allow to distinguish between the different actors, making it difficult, if not impossible, to interpret the story correctly. For example,  in our example, if both ``María Rodríguez'' and ``Juana Fernández`` are replaced by  an X, it would be difficult to interpret that the lady that made the complaint is not the  same as the one  that sold the phone chip.

To avoid this problem, mentions are replaced by  fantasy names  or just a generic label,  associated in a unequivocal way to each actor in the text. This last method is used, for example, in our corpus where the labels are fictitious initials: AA, BB, etc. In our previous example,  mentions ``Mr. Juan Pérez'' and ``Mr. Perez'' are replaced by ``Mr. AA'' since both refer to the same person.

In return, the process of de-identification is now more complex: it is no longer enough to detect and suppress all names; consistency must be maintained in assigning the new labels to the original actors, which implies the resolution of coreferences, at least between the different variants of proper names.

This task presents its difficulties. For example, consider the following  excerpts taken from a sentence of the Family Court of Appeals:

	\begin{enumerate} \label{ej:sent99-2009}
			\item \label{ej:sent99-2009-a}{\small Jorge Martínez, Juan Líber c/ Pérez Rodríguez, Pedro y otros. }  \\  
														 {\tiny Jorge Martínez, Juan Líber against Pérez Rodríguez, Pedro and others}
			\item \label{ej:sent99-2009-b}{\small (\ldots) Sres. \textbf{Pedro y Juan Pérez}, deduce recursos de apelación.  } \\ 
											{\tiny (\ldots) Misters Pedro y Juan Pérez, deduct appeals. }
			\item \label{ej:sent99-2009-c}{\small No puede considerarse que \textbf{Pedro Pérez} ha omitido contestar la demanda(\ldots)} \\
														{\tiny It can not be considered that Pedro Pérez has omitted to answer the demand(\ldots) }           
			\item \label{ej:sent99-2009-d}{\small (\ldots) el Sr. \textbf{Pedro Pérez} por haber precluido la oportunidad procesal para promoverla(\ldots) }      \\ 
														{\tiny (\ldots) Mr. Pedro Pérez for having precluded the procedural opportunity to promote it (\ldots)}
			\item \label{ej:sent99-2009-e} {\small(\ldots) en donde \textbf{Pedro Pérez Rodríguez} denunció que su domicilio era (\ldots)}       \\
														{\tiny (\ldots) in which Pedro Pérez Rodríguez denounced that his domicile was (\ldots)}       
			\item \label{ej:sent99-2009-f}{\small Se intimó la aceptación de \textbf{Pedro} a fs. 32 vta. y a \textbf{Juan} a fs. 36/37 (\ldots)}   \\
														{\tiny It was asked the acceptance  of Pedro in fs. 32 vta. and  Juan in fs. 36/37 (\ldots)}
	\end{enumerate}

It must be taken into account that, for the substitution process, ``Pérez Rodríguez, Pedro'', ``Pedro'' and ``Pedro Pérez Rodríguez'' (fragments \ref{ej:sent99-2009-a}, \ref{ej:sent99-2009-b}, \ref{ej:sent99-2009-e} and \ref{ej:sent99-2009-f}) refer to the same person,  ``Juan Pérez'' and ``Juan'' (fragments \ref{ej:sent99-2009-f} and \ref{ej:sent99-2009-b}) refer to another. In particular, from the fragment \ref{ej:sent99-2009-b} it is inferred,  from the plural ``Misters'' and the distribution property of the conjunction \emph{and},  that the aforementioned ``Pedro'' has ``Pérez'' as surname, while in the fragment \ref{ej:sent99-2009-f} it must be assumed that ``Pedro'' and ``Juan'' are the same ``Pérez'' brothers because they are the only ones mentioned with those given names in that piece of text.

\section{Corpus}
Our corpus consists of a part of the National Jurisprudence Base, composed of circa \numprint{79000} documents, of which about \numprint{17000} are manually de-identified. This gives us,  a priori, a parallel corpus to work with, from which we can, for example, train a machine learning model or evaluate existing tools.  Unluckily, many of these \numprint{17000}  documents have failures in their anonymization process:  there are names left out or  labels incorrectly assigned to actors .

With the purpose of establishing a gold standard  we proceeded to the revision of \numprint{1000} documents, identifying and labeling manually all the entities to be de-identified with the aid of  BRAT \cite{BRAT12}. A total of \numprint{10102} entities distributed in 997 documents are manually tagged, while 3 documents are finally discarded because they do not contain entities to de-identify. We use these \numprint{1000} documents for evaluating state of the art tools: Freeling \cite{padro12}, CoreNLP \cite{manning14}, Nltk \cite{bird09} and Spacy \cite{Honnibal17}.

All tested NLP tools present issues processing our corpus due to its characteristics which differentiates them from journalistic or scientific texts (the usual kind of text these tools are trained with).  The problems detected appear in the early stages of what is a standard text analysis: the analyzers make important errors, from  tokenization and  segmentation to  detection of named entities --particularly important for our problem--. For example, the difference in number of sentences and tokens detected by Core NLP and Freeling in the whole corpus is appreciable: close to \numprint{100000} for the first, \numprint{500000} for the second.

Table \ref{corpus:cuadro:comparativo:segmentador}  contains examples of sentences incorrectly marked by three different analyzers, NltK, Freeling and CoreNLP, over the same set of documents. From a quantity perspective, the number of sentences detected is quite different for each analyzer: Nltk detects 136 sentences, while CoreNLP finds 87 and Freeling  detects only 47.  From the analysis of the results,  we detect that periods used in abbreviations or ordinals are incorrectly managed  like with "7o."\footnote{Ordinal number,  "7th.".}  (sentence 44), "Sr."\footnote{Abbreviation of "Señor", "Mister".} (sentence 56) or  "pág."\footnote{Abbreviation of "página",  "page".}  (sentence 108).

\begin{table} 
	\begin{center}
		\setlength{\tabcolsep}{5pt}
		\begin{tabular}{|p{\linewidth}|}
			\hline	
			Freeling:\\
			\hline
			Sentence 44:  En el mismo sentido se pronunció el Tribunal en lo Civil de 7o. \\
			Sentence 45:  Turno, cuando en Sentencia No . 134/98 sostuvo que...\\
			\hline		
			CoreNLP: \\
			\hline
			Sentence 56:  Como lo plantea el Sr.  \\
			Sentence 57:  Fiscal de Corte a fs . \\
			Sentence 58:  183 v., si bien ateniéndonos ...\\
			\hline
			Nltk:\\
			\hline
			Sentence 108:  (Nicolás Coviello, Doctrina General del Derecho Civil, pág.\\
			Sentence 109:  78, Cf.\\
			Sentence 110:  Eduardo García Maynez, Introducción al estudio del derecho, pág.\\
			Sentence 111:  329)" \\
			\hline		   
		\end{tabular}
	\end{center}

	\caption{Examples of incorrect sentence boundary recognition by Nltk, CoreNLP and Freeling.  Each group (108-111, 56-58 and 44-45) should belong to the same sentence even tough they are split by the segmenters.} \label{corpus:cuadro:comparativo:segmentador}
		
\end{table}

\begin{figure}
	\begin{center}
		\includegraphics[width=20em]{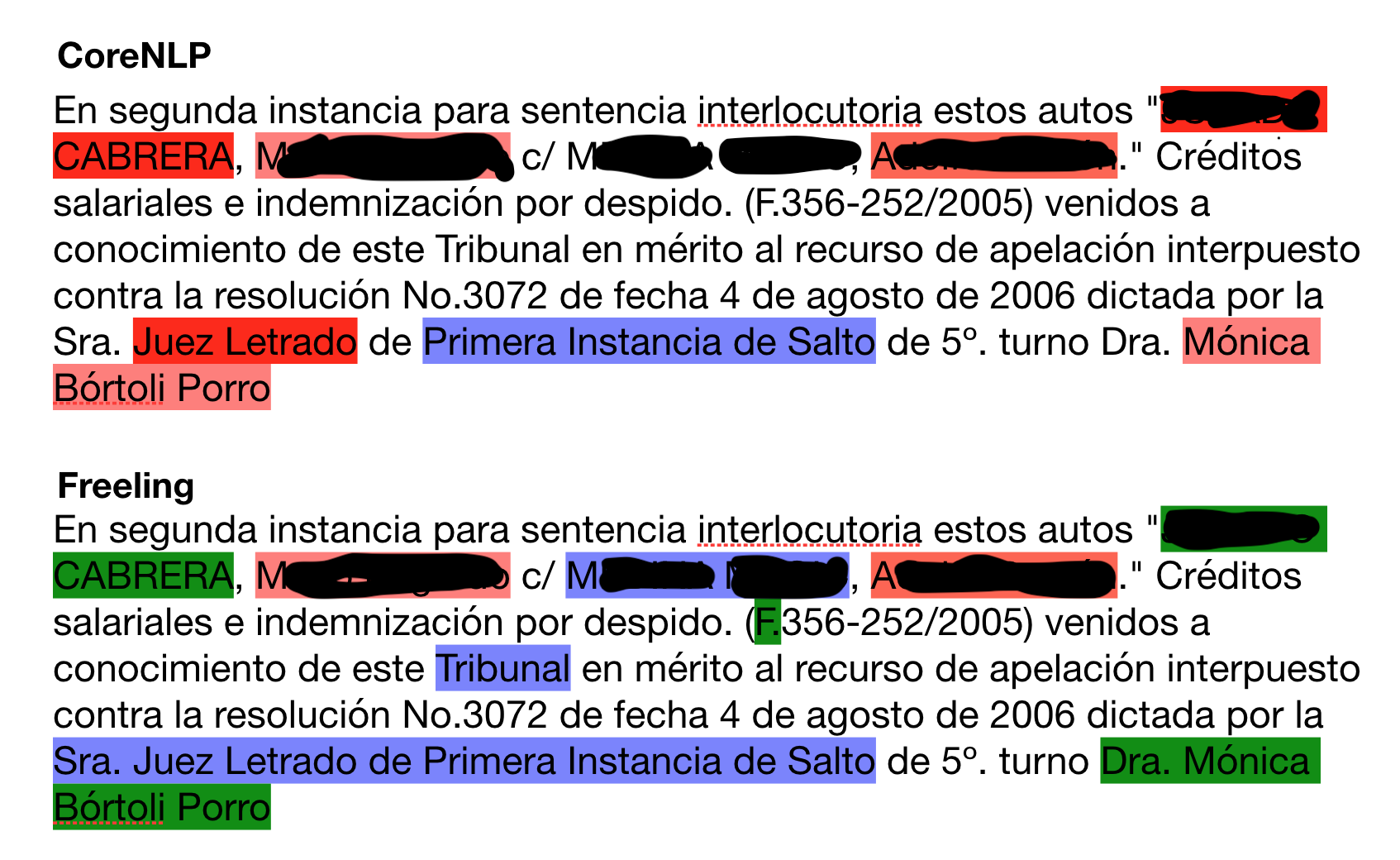} 
	\end{center}
	\caption{NER by CoreNLP and Freeling; highlighted words are NER entities (color red for persons) . Both NER fail to recognize "XX Cabrera, XX" and " Mx XX, Axx" as two references of persons at the beginning of the text.}
	\label{corpus:fig:ner}
\end{figure}

There are also problems in the detection of named entities by all three tools. Figure \ref{corpus:fig:ner} shows the result of applying the NER modules of  Freeling and CoreNLP, where it can be seen that not only are the segments incompletely recognized, but also, there are problems with the types assigned to the different entities.

\begin{table} 
	\begin{center}
		\setlength{\tabcolsep}{5pt}
		\begin{tabular}{|l|r|r|r|}
			\hline
			& Freeling & Spacy & CoreNLP 	\\
			\hline
			Marked ent.	&  49352	&	78304 & 112330\\ 
			\hline
			Perfect ent.	& 	7367	&	7092 & 4167\\
			\hline
			Covered ent.	& 	8448 	 &	8336 & 4461 \\
			\hline
			Perfect doc.  & 	121	&	82 & 7\\
			\hline
			Covered  doc.	& 	  160	 &	118 & 10 \\
			\hline					    							   
		\end{tabular}
		
	\end{center}
	\caption{Entities detected by  Freeling, Spacy and CoreNLP. Only a fraction of our goal (10102 entites) is found, even when tools recover more than \numprint{50000} entities of several types. } \label{corpus:cuadro:comparativo}
\end{table}

In the understanding that the priority in de-identification is  hiding of all sensitive names, the recall is measured according to three tools: Freeling, Spacy and CoreNLP. Two cases are considered: when the entity is detected by the analyzer completely, with its correct limits (\emph{perfect match}), and a second case in which the text segment covers the whole entity, but includes more text (\emph{obscure match}).

 The results of the three analyzers are shown in Table \ref{corpus:cuadro:comparativo}. From all 997 documents, only  1\%-16\%  of the documents get all its entities detected or completely covered\footnote{An name is completely covered if a tagged entity comprises all of the name's tokens but possibly more. }, and just 7\%-12\% get all of their entities recognized. That is, even considering the most lenient measure, 84\% of the sentences would have at least one name partially left as-is, risking the re-identification of people involved in those texts. Therefore, state of the art tools cannot deal with these kind of text without adjusting and retraining of their different modules.

\section{Measures}
For the evaluation of the anonymization systems, different measures have been proposed depending on the final goal of the task. Some of these measures have their origin in Information Retrieval (IR) --those regarding quantity and quality of the marked entities--, while others come from the Resolution of Coreferences \cite{DIAS16} --those that seek to maintain a unequivocal link between labels and original names--.

From IR come Precision and Recall and $F_1$\cite{MANNING99}, in their  micro and macro versions of these measures: for the first one each instance counts as one entity, for the second, each document counts as one. 
 
But Precision and Recall  do not take into account if all the references to the same real entity are recognized as such, that is, if the coreference chains are correctly solved by the system. Specific metrics are proposed for coreference resolution, where each entity is seen as a set of linked mentions and it is evaluated how close is the partition detected by the system of the real sets of entities present in the text.

Analogous to the micro and macro versions of $ P $ and $ R $, in the measures for the resolution of coreferences, a distinction is made between those based on links --each mention in the text has equal weight, with which entities with more references weight more than others with fewer-- and those based on instances --each entity has equal weight regardless of the number of references that occur in the text--.

Although there are several measures proposed,  there is no consensus on which is the best. Among the first ones, which arise in the 90s, are MUC and $ B^3 $ \cite{BAGGA98, VILAIN1995}, although, due to the weaknesses they present, several alternatives are proposed such as CEAF, BLANC and LEA \cite{LUO05, MARQUEZ13, MOOSAVI16}.

For example, consider $S$ as the real set of two entities, where each entity is represented by the set of mentions in the text; and  $T$ as the output partition given by the system:

\begin{center}
$  S= \{ \{a_1, a_2, a_3\}, \{b_1, b_2, b_3, b_4\}  \}$    \hspace{1em} 
$ T=\{ \{a_1, a_2\}, \{a_3, b_1\}, \{b_3, b_4, c_1\}  \}$
\end{center}

Table \ref{medidas:cuadro:comparativo} shows the result of several measures for the previous  example. As it can be observed, the values  for the problem posed as an example can be quite different, varying from a minimum of 0.31 to a maximum of 0.57. 

To our knowledge, it does not exist a global measure for the anonymization task.

\begin{table} 
	\begin{center}
		\setlength{\tabcolsep}{5pt}
		\begin{tabular}{|l|c|c|c|}
			\hline
			& P & R & F 	\\
			\hline
			$MUC$					& 	0.40	&	0.50 & 0.44\\
			\hline
			$B^3$					  & 	0.42	&	0.62 & 0.50\\
			\hline
			$CEAF_{\Phi_3}$		& 	  0.57	 &	0.57 & 0.57 \\
			\hline
			$CEAF_{\Phi_4}$		& 	  0.69	 &	0.46 & 0.55 \\
			\hline
			$BLANC$					& 	  --	 &	--  	& 0.43 \\
			\hline
			$LEA$						& 	  0.24	 &	0.34 & 0.31 \\
			\hline						    							    
		\end{tabular}
		
	\end{center}
	\caption{$P$, $R$ y $F$ for several coreference evaluation measures over the same example.} \label{medidas:cuadro:comparativo}
\end{table}

\section{Conclusions}
The first stages of an anonymization project of sensitive data in legal  documents are presented and discussed, in a context where the interest in making the information of the sentences publicly available collides with the right to privacy of people involved in the trials. Given a parallel corpus of \numprint{17000} sentences, our first thought was to apply supervised learning algorithms with this data. But an analysis of the performance of the available  tools for some basic needed preprocessing  --tokenization, sentence segmentation, named entities recognition-- shows the in-feasibility of this option and the need to retrain these tools in order to  encompass the legal domain. In addition to the insufficiency of usual NLP tools, it is also necessary to complement the existing theoretical developments in terms of joint evaluation measures that consider at the same time  the  identification of entities in texts and the construction of coreference chains.  A reflection that arises is if training and evaluation of NLP tools do not correspond, in fact, to somewhat simplified data,  without enough variety to allow a more fluid passage to real domains of application.

\bibliography{biblio}
\bibliographystyle{acl_natbib}

\end{document}